\documentclass{article}
\usepackage{spconf,amsmath,graphicx}
\usepackage{xspace}
\usepackage{amsfonts}
\usepackage{times}
\usepackage{color}
\usepackage{hyperref}
\usepackage{url}
\usepackage{alltt}
\usepackage{tabularx}
\usepackage{tikz}
\usepackage{multirow}

\usepackage{float}
\usepackage{todonotes}
\usepackage{enumitem}

\setlist{nosep, leftmargin=14pt}

\usepackage{mwe} % to get dummy images

\title{3D Vertebrae Measurements: Assessing Vertebral Dimensions in Human Spine Mesh Models Using Local Anatomical Vertebral Axes}
\name{Ivanna Kramer, Vinzent Rittel, Lara Blomenkamp, Sabine Bauer, Dietrich Paulus}
\address{Institute for Computational Visualistics, University Koblenz, 56070 Koblenz, Germany}
\begin{document}
%\ninept
%
\maketitle
\begin{abstract}
Vertebral morphological measurements are important across various disciplines, including spinal biomechanics and clinical applications, pre- and post-operatively. These measurements also play a crucial role in anthropological longitudinal studies, where spinal metrics are repeatedly documented over extended periods. Traditionally, such measurements have been manually conducted, a process that is time-consuming. In this study, we introduce a novel, fully automated method for measuring vertebral morphology using 3D meshes of lumbar and thoracic spine models.Our experimental results demonstrate the method's capability to accurately measure low-resolution patient-specific vertebral meshes with mean absolute error (MAE) of 1.09 mm and those derived from artificially created lumbar spines, where the average MAE value was 0.7 mm. Our qualitative analysis indicates that measurements obtained using our method on 3D spine models can be accurately reprojected back onto the original medical images if these images are available.
\end{abstract}
\begin{keywords}
morphological measurements, spine meshes, vertebrae measurements, spinal CT
\end{keywords}
\section{Introduction}
\label{sec:intro}

% -------------------------------------------------------------------------
The quantitative assessment of vertebral morphological parameters encompasses the measurement of various dimensions and characteristics of spinal vertebrae. These dimensional measurements are pivotal in diverse domains such as research, clinical practice, and anthropology, aiding in the understanding of patient-specific anatomical variations and the analysis of pathologies affecting the vertebral column. Typically, in vivo these measurements are derived from lateral and transversal cross-sectional images obtained via CT, MRI, or X-ray imaging techniques \cite{Davies2009NVD,Zhou2000GDO,Singh2021CTM, Wu2021MAO}. In clinical settings, routinely measured parameters include vertebral height, width, and depth. These patient-specific morphometric data are crucial for preoperative planning and medical interventions. They are especially pertinent in scenarios requiring patient-specific morphological data for implant design and placement, as well as in the assessment of fractures.

%To measure these parameters in vivo, traditionally manual measuring protocols involves  sagittal radiographs  \cite{FROBIN1997S1} or magnetic resonance images \cite{COPPOCK2023100378} and computer tomography \cite{flanders2023height}.

\begin{figure}[htb]
 \centering
  \centerline{\includegraphics[width=8.5cm]{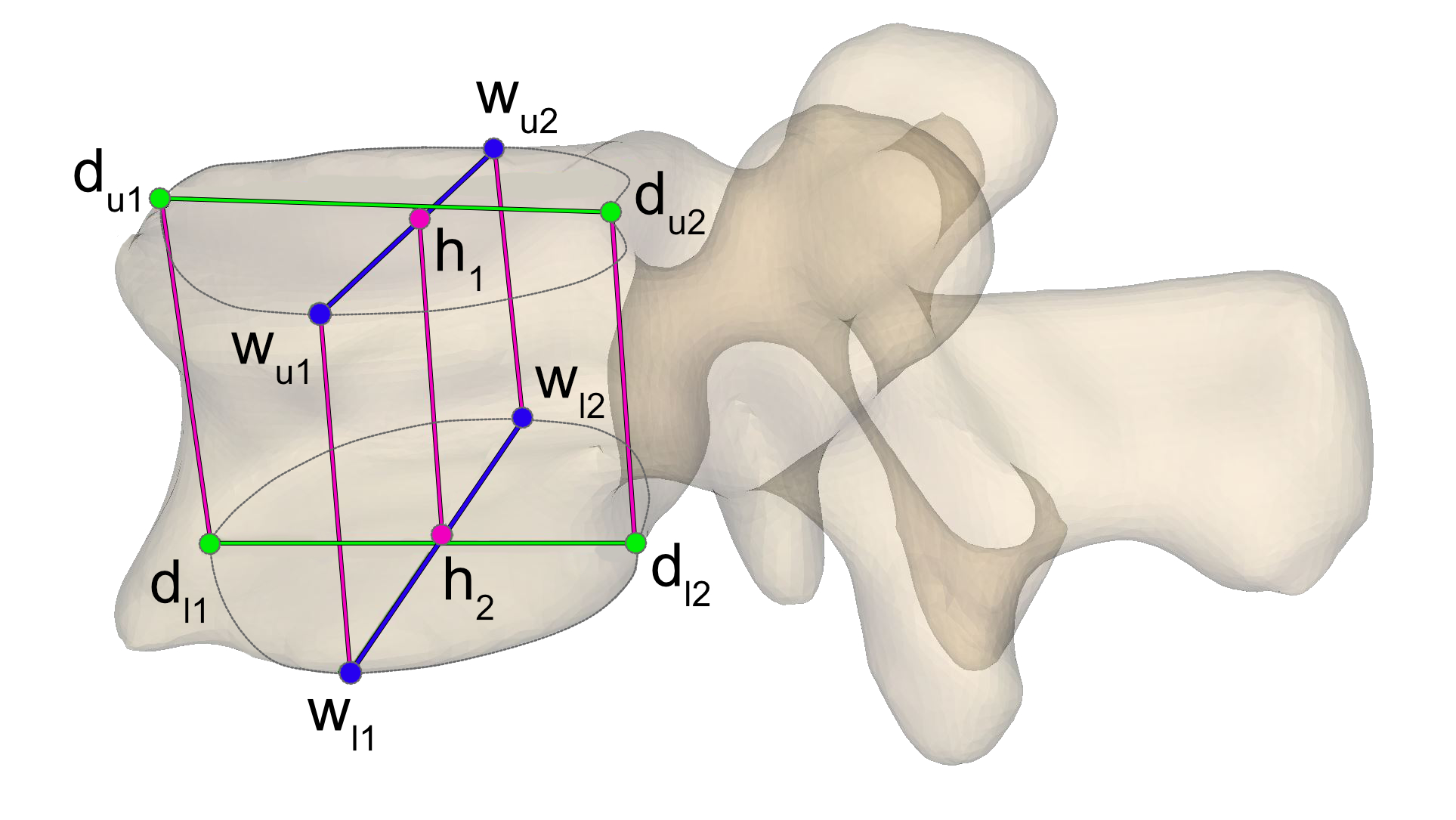}}
\caption{Sample measurements of established dimensional parameters of the vertebral body shown on a 3D model of L2}  
\label{fig:sample_measurements}
\end{figure}

Vertebral fractures often do not present as a clinically recognizable event. Among patients over 60 years of age presenting to emergency departments, approximately 1/6 had a moderate to severe vertebral fracture evident on lateral chest radiographs, of which only about 1/2 were noted on radiology reports and even fewer received specific medical attention \cite{Griffith2015IOV}.  Therefore a specific application of our method could be the measurement of vertebral body height over a longer period of time to detect a decrease in vertebral body height due to degenerative processes such as osteoporosis at an early stage. Our method would allow medical measures to be taken in time before a complete collapse of the vertebral body occurs. 

In this paper, our contribution is threefold: we present a novel method that can accurately measure vertebral body dimensions on the 3D spinal vertebral meshes in a fully automated manner; we extend a part of an existing publicly available spine dataset with the morphological measurements. Finally, we publish an artificially created spine dataset together with annotations for the reproducibility of the conducted experiments \footnote{\url{https://github.com/VisSim-UniKO/Spine-Morphology-Analyzer}}. We also present the implementation of the method as 3D Slicer \cite{pieper20043d} plugin that is ready to use for measurements of lumbar and thoracic 3D spinal models.

\section{Recent work}
\label{sec:recent_work}

In the publication of Tan et al. \cite{tan2012high}, they propose a semi-automatic method to measure the height of the vertebral bodies from spinal CT images. Their approach follows several step including CT segmentation and 3D mesh generation. On the segmented vertebrae endplates, evolving level set algorithms are performed. Finally, the vertebral body height is calculated as the average distance between all lines connecting counterpart points on the opposite endplates. The method was tested on scanned and segmented phantoms and showed a performance up to four times more accurate than manual measuring. Angelo et al. \cite{di2015new} published a fully automated method for determining the vertebrae dimensions. A method based on mirroring and registration was first applied to identify multiple geometrical planes that divide the vertebra model into coronal, frontal, and sagittal planes. The calculation of the specific dimensional features is performed on the intersections of the geometrical planes with the corresponding parts of vertebrae. The conducted experiments have shown, that the accuracy of the identification of the geometrical measurements depends on the point density in the meshes: it was stated that the error increases for those meshes that are lower than six points per $mm²$. Alukaev et al. \cite{alukaev2022deep} proposed the automated approach utilizing an  ensemble of  U-Net based neural networks to estimate vertebral morphometry including measurements for the height of the vertebral body and heights of the intervertebral discs in the CT images of the spine. 

\section{Materials and Methods}
\label{sec:pagestyle}

\subsection{Measured dimensions}
\label{sec:dimenstions}
We examine vertebra body's key parameters (see Fig. \ref{fig:sample_measurements}), critical in biomechanics, anthropology and forensic studies: $w_{u1}w_{u2}$ being the  width of the superior vertebral body's endplate, $w_{l1}w_{l2}$ being the width of the inferior one. The depth is defined as line  $d_{u1}d_{u2}$ and $d_{l1}d_{l2}$ for superior and inferior endplates, respectively. The five measured vertebra heights are: anterior height $d_{u1}d_{l1}$, central height $h_{1}h_{2}$, posterior height $d_{u2}d_{l2}$, left $w_{u1}w_{l1}$ and right $w_{u2}w_{l2}$ height. All parameters are measured in \textit{mm}.  

\subsection{Approach}

\begin{figure*}[h!]
  \centering
  \includegraphics[width=17.5cm]{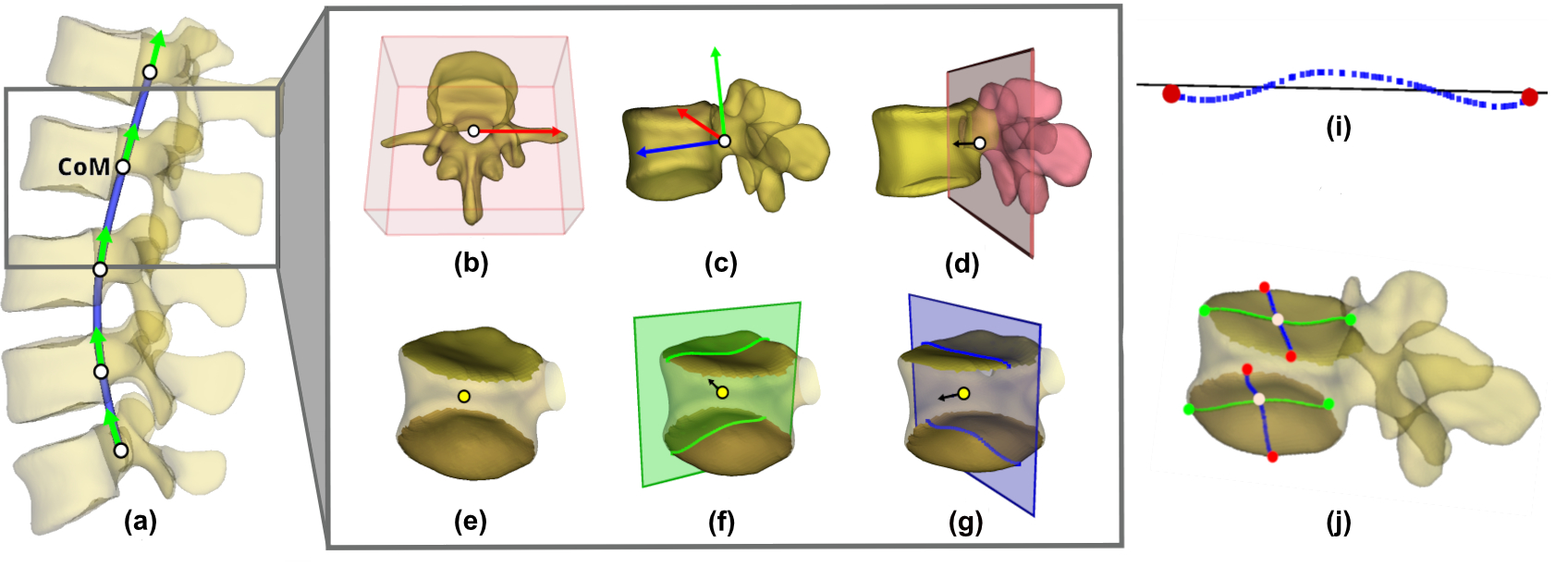}
\caption{Overview of the proposed approach: the local vertebrae orientations are calculated first(\textbf{(a)}-\textbf{(c)}). Subsequently, cutting planes in the frontal view (\textbf{d}) are established based on the Center of Mass (CoM) and a normal vector derived from step \textbf{(c)}. This procedure precisely segments the vertebral geometry into focused region of interest, i.e. vertebral body. In the next step, the upper and lower endplates are extracted from the vertebral body (\textbf{(e)}). In final steps, the intersection points are calculated after cutting the vertebral body with the local frontal and sagittal planes. The landmarks (marked red and green in \textbf{(i-j)}) for the dimensional measurements then derived from the these intersections and are used to identify the vertebral dimensions.}
\label{fig:approach}
\end{figure*}

The proposed method for automated measurements of the vertebral dimensional characteristics is performed in several steps, which is depicted in Fig. \ref{fig:approach}. Based on local orientations of the vertebrae geometry in the spine model, we define anatomical planes, that are used to identify the vertebral bodies.  Once these bodies are identified, we proceed to measure their dimensions. The main assumption in our method is, that the geometries follow the Left-Posterior-Superior (LPS) coordinate system. 

\textbf{Local vertebra axes}: The initial step of our proposed method involves determining the center of mass (CoM) for each vertebra in the spine, denoted by a white dot in Fig. \ref{fig:approach} (a). The CoM is defined as the average position of all surface vertices of the vertebra. Once, all CoMs are determined, we fit a cubic spline, which facilitates a smooth curve, through all CoM points. The local up-vector for each vertebra, depicted in green in Fig. \ref{fig:approach} (a), is then computed as the first derivative of this spline curve at each CoM location.

The subsequent step involves calculating object-oriented bounding boxes of each vertebra to approximate the principal orientations of the mesh. Due to our previous observations, we find the only accurate, usable orientation from the bounding box is the vector that defines the right local axis of the individual vertebrae. In order to identify each local right vector ($\Vec{v}_{right}$), we maximize the dot product between the local orientation vector provided by the bounding box and the global right axis. The vector that yields the maximum dot product represents the local right-vector, see Fig. \ref{fig:approach} (b) for each vertebra. Lastly, we determine the local frontal axis of the vertebra model. The cross product between the local up- and right vectors is equal to the local front axis ($\Vec{v}_{front}$), depicted in blue in Fig. \ref{fig:approach} (c). 

\textbf{Region of interest}: Our focus is to measure geometrical dimensions of the vertebral body, keeping only those vertebral parts that are relevant to our measurements. This involves a segmenting the vertebrae into their anterior and posterior components. Therefore, we establish a cutting plane defined using the vertebra's local sagittal plane, the CoM position calculated in the previous step, and the normal to the plane $\Vec{v}_{sag}$ pointing to the front.

After the vertebra mesh is divided, the portions of the geometry not contributing to the dimensional measurements is trimmed away, as illustrated in Fig. \ref{fig:approach} (d). Simultaneously, we update the remaining mesh's CoM ( white dot in Fig. \ref{fig:approach} (e)).

\textbf{Extraction of the endplates:} After identifying the vertebral body as the region of interest, we focus on extracting its upper and lower endplates. This extraction is based on analyzing the normal vectors ($ \Vec{n}_p \in \mathbb{R}^3$) for each  vertex $p$ on the mesh. To segment the vertebral endplates, we first identify and remove vertices whose normal vectors significantly deviate from the local up-axis $\Vec{v}_{up}$ of the vertebral body. This deviation is quantified by projecting each vertex's normal vector onto $\Vec{v}_{up}$, calculated as $proj(p_i) = n_i \cdot \Vec{v}_{up}$. We then define a specific threshold, $max\_cos$ and $-max\_cos$, to differentiate vertices that belong to the upper and lower endplates correspondingly. Vertices are considered part of the upper endplate if their projection value is less than $max\_cos$, and part of the lower endplate if their projection value is greater than $-max\_cos$. We found the value for $|max\_cos|$ of $\deg{45}$ aligns perfectly with our experimental results. The segmented endplates are visualized as solid, non-transparent meshes in Figure \ref{fig:approach} (e).

\textbf{Landmarks determination and dimensions calculation:} In the last step of the approach, we calculate specific landmarks, to determine the dimensions outlined in Sec. \ref{sec:dimenstions}. ja
This involves calculating the intersections of the vertebral endplates with the local sagittal and frontal planes. These planes are defined by the respective local axes, acting as normals to the planes, and the center of mass of the segmented vertebral body, illustrated by the green and blue planes in Fig. \ref{fig:approach} (g)-(e).  To identify the required landmarks, we arrange the intersection points along their respective axes and identify the minimum and maximum points as landmarks (as shown in Fig. \ref{fig:approach} (i)-(j)). Finally, the Euclidean distances between these landmarks are calculated to obtain the required dimensional measurements. 

\subsection{Datasets}
To establish the reliability and generalizability of the proposed method, two distinct datasets containing different 3D spine models were employed. 

\begin{table*}[h!]
\centering
\small\addtolength{\tabcolsep}{0pt}
\begin{tabular}{|l| c| c| c| c| c| c| c| c| c| c|} 
\hline
\textbf{Dataset} & \textbf{Spine}  &  $\mathbf{w_{u1}w_{u2}}$ & $\mathbf{w_{l1}w_{l2}}$ & $\mathbf{d_{u1}d_{u2}}$ &$\mathbf{d_{l1}d_{l2}}$ &$\mathbf{h_{1}h_{2}}$ &$\mathbf{d_{u1}d_{l1}}$ &$\mathbf{d_{u2}d_{l2}}$ & $\mathbf{w_{u1}w_{l1}}$ & $\mathbf{w_{u2}w_{l2}}$ \\ %[0.5ex] 
\hline \hline
    \multirow{2}{*}{VerSe2020 MAE}
        & lumbar  &  0.75	& 0.88 &	0.49 &	0.54	& 0.39     &	1.32 &	0.69 &	1.10 &	0.75  \\
    \cline{2-11}
        &  thoracic & 3.73	 & 0.26	 &1.22	 & 1.45	 & 0.37	  & 	7.64 &  1.27 & 0.72 & 2.65\\
    \cline{2-11}
\hline
\hline
  \multirow{4}{*}{VerSe2020 ICC}
     & lumbar &  0.93 &0.98 &0.93 &0.98 &0.94 & 0.52 &0.48 &0.94 &0.89   \\
    \cline{2-11}
    & lumbar* &  0.75	& 0.88 &	0.49 &	0.54 & 0.39   &	1.32 &	0.69 &	1.10 &	0.75  \\
    \cline{2-11}
    \cline{2-11}
     &  thoracic & 0.98 &0.98 &0.92 & 0.99 & 0.99 & 0.98 &0.94 & 0.97& 0.95\\
    \cline{2-11}
     &  thoracic* & 0.97 &0.98 &0.93 &0.97 &0.99 &0.63 &0.96  & 0.89 &0.91\\
\hline
\hline

\multirow{1}{*}{\shortstack[l]{Sawbone (MAE)}}
        & L1-L5 & 0.89&0.86&0.73&0.80&0.45&0.35&0.12&1.25&0.78 \\
         \cline{2-11}         
\hline
\end{tabular}
\caption{ Mean Absolute Error (MAE) [mm] and Interclass Correlation (ICC) Across Vertebral Segments and Dimensional Parameters. The ICC values are presented separately for human raters (rows without *) and for a combined assessment of both raters and measurements predicted by our method (rows with *).}
\label{tab:results_short}
\vspace{-0.3cm}
\end{table*}

\subsubsection{VerSe}
The VerSe dataset \cite{loffler2020vertebral} comprises CT images and corresponding segmented spines from a total of 141 patients. To validate our method, we randomly selected six patients from this dataset and extract the vertebrae segments as 3D meshes. Dimensional parameters, as outlined in Fig.\ref{fig:sample_measurements}, were measured on these patients by three independent raters using 3D Slicer. Specifically, the L2, L3, T4 and T5 vertebrae were chosen for measurements in each selected patient. Considering our focus on 3D mesh data, we also examined the mesh density of the geometries provided in the VerSe dataset. The average density of the geometries in the dataste was 0.16 vertices per square millimeter (0.16/mm²), indicating a relatively low resolution of 3D meshes. To assess the consistency of the measurements made by the raters, we employed the Intraclass Correlation Coefficient (ICC) \cite{shrout1979intraclass} for each dimensional parameter measured. The high average ICC value of 0.84 for lumbar  and 0.96 for thoracic vertebrae signify a strong reliability in the measurements between the experts. However, the variability in manual measurements of the width, with a standard deviation of 3.41 mm, highlighted this dimension as a challenging area for raters, possibly due to complex anatomical features in some fractured vertebrae. 

\subsubsection{Artificial Sawbone Dataset}

Another dataset, that was used for validation, was generated on a set of Sawbone meshes \cite{zdero2023biomechanical} by following the approach described in \cite{10278898}. The dataset comprises a collection of 50 artificial generated lumbar spine models and the corresponding dimensional measurements. The lordosis angles for these generated spine models were randomly selected, spanning a range from 40\textdegree~ to 74\textdegree. To ensure the wide spectrum of dimensional variations within the dataset, the vertebrae in the models underwent random scaling. This scaling process adjusted the vertebral bodies' dimensions to fit within specified intervals: the vertebral width varies between 34.84 mm and 62.78 mm, the depth ranges from 24.20 mm to 45.10 mm, and the vertebral body height, measured at the midpoint, falls between 16.84 mm and 33.16 mm. 

\section{Experiments and results}
\label{sec:typestyle}

We evaluated the proposed method using two different metrics. We applied mean absolute error (MAE) with $ MAE =\frac{1}{n}\sum_{i=1}^{n}|p_i - t_i|$  between the $n$ averaged over all raters ground truth measurements $t_i$ and the corresponding predicted values $p_i$. Additionally, we calculated ICC between raters and the automated approach, to provide a measure for consistency value between human experts and the automated system. The results including MAE and ICC values for two distinct datasets are presented in Table \ref{tab:results_short}. On the VerSe dataset, the average deviation of the proposed method over all vertebral dimensions from the raters' mean measurements was 1.32 mm for lumbar spine, suggesting that the approach generally aligns closely with the central tendency of rater assessments. A relative large deviation of 7.64 mm to the ground truth value was shown in particular for the anterior height measurements $d_{u_{1}}d_{l_1}$ in thoracic vertebrae, which also reflects in the moderate ICC of 0.63. 

The validation dataset included measurements of two abnormal vertebrae, which provided a stringent test of the system's robustness. The proposed method exhibited minimal error in these edge cases, with deviations from the rater's measurements remaining under 1.34 mm for the L2 vertebra and under 1.15 mm for all dimensions except the lower endplate's depth (1.15 mm) for the L3 vertebra. These results underscore the method's capability to handle atypical anatomical variations with a high degree of precision.  On the synthetic Sawbone dataset our method showed MAE smaller than 1.25 mm for the height measurements, with an overall average MAE of 0.7 mm.  The average MAE values across vertebrae are close to each other, suggesting a relatively consistent performance across different spinal levels. 

\begin{figure}[htb]
 \centering
  \centerline{\includegraphics[width=6.5cm, height=5.5cm]{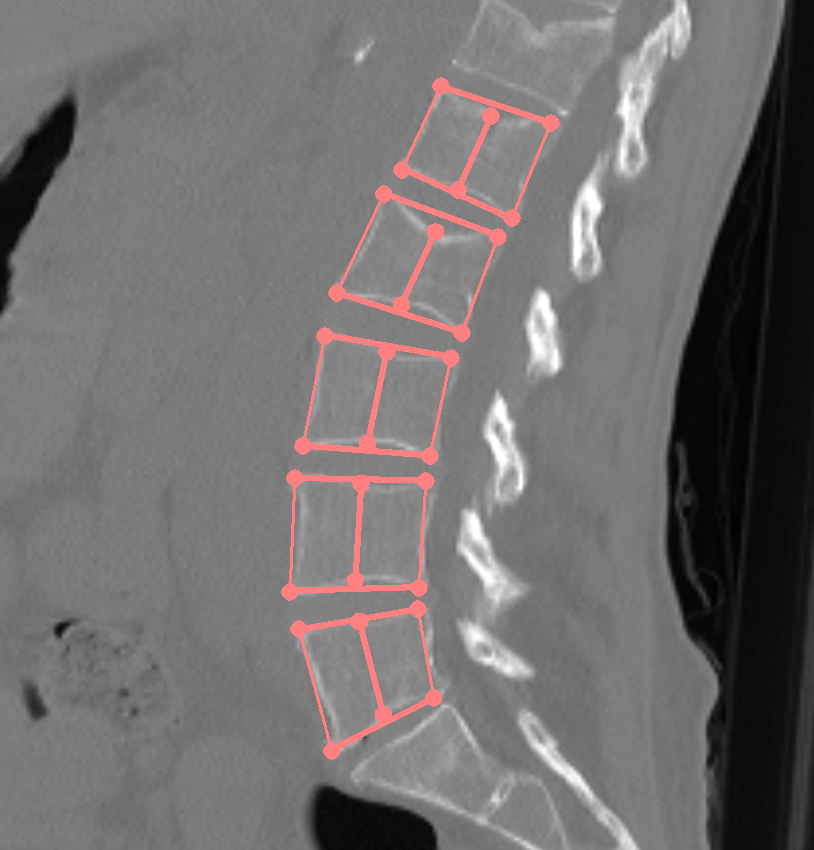}}
\caption{Back-projection of vertebral measurements into its corresponding sagittal CT scan of VerSe2020 dataset. Six landmarks in the local sagittal plane define the outer distances as well as the vertebral body height in the center.}  
\label{fig:plugin-overlay}
\end{figure}

Qualitative results are presented in Fig. \ref{fig:plugin-overlay}. Our method successfully reprojects the vertebral dimensions measured on 3D spine meshes back onto CT scans. Furthermore, this add-on allows the utilization of automatically determined landmarks in the original medical images possibly enhancing the precision of the measurements.
Nonetheless, additional experiments and validation are necessary to confirm the efficacy of the above mentioned reprojection feature.

\section{Discussion and conclusion}
\label{sec:conclusion}

In this study, we introduced an automated method for measuring spine anthropomorphic parameters in 3D meshes, which has shown comparable accuracy to experts even when assessing vertebrae with structures. In comparison with \cite{di2015new}, that operate on the highly dense meshes with 15 points on mm$^2$, our approach can provide accurate measurements on the low-resolution meshes. While our results are promising, indicating the tool's reliability and consistency, the variability observed in manual measurements show a need for further validation. Our tool has potential applications in improving measurement efficiency in both clinical and research domains. Future work will expand the system's capabilities to include rotational parameters and cervical vertebrae measurements, and to contribute to the development of algorithms for vertebral morphometry.

\section{Compliance with Ethical Standards}
This research study was conducted retrospectively using human subject data made available in open access. Ethical approval was *not* required as confirmed by the license attached with the open access data.

\bibliographystyle{IEEEbib}
\bibliography{strings,refs}

\end{document}